# Spatial Fuzzy C-Means PET Image Segmentation of Neurodegenerative Disorder

A. Meena

Research Scholar, Dept. of CSE,
Sathyabama University, Chennai
Tamilnadu, India
kabimeena2@hotmail.com

K. Raja

Dean(Academics), Alpha College of Engineering ,
Chennai, Tamilnadu, India
raja_koth@yahoo.co.in

**Abstract**

Nuclear image has emerged as a promising research work in medical field. Images from different modality meet its own challenge. Positron Emission Tomography (PET) image may help to precisely localize disease to assist in planning the right treatment for each case and saving valuable time. In this paper, a novel approach of Spatial Fuzzy C-Means (PET-SFCM) clustering algorithm is introduced on PET scan image datasets. The proposed algorithm is incorporated the spatial neighborhood information with traditional FCM and updating the objective function of each cluster. This algorithm is implemented and tested on huge data collection of patients with brain neuro degenerative disorder such as Alzheimer's disease. It has demonstrated its effectiveness by testing it for real world patient data sets. Experimental results are compared with conventional FCM and K-Means clustering algorithm. The performance of the PET-SFCM provides satisfactory results compared with other two algorithms.

## 1. Introduction

Image diagnosis is a major component for treatment planning, research activities and more. Currently, the algorithm computation is playing an important role due to the increasing size and number of medical images. The selection of methods mainly depends on imaging modalities, its specific application and other factors. For an example the brain tissue has different requirements from other organ. Medical image segmentation automates the specific radiological function and other regions of interest. The goal of segmentation is to analyze the representation of an image into meaningful and easier. It refers to partition a digital image into multiple segments which are basically constructs with sets of pixels. Each pixel in the region of interest consists of some basic characteristics and computed property, known as intensity, texture and color. But, no single segmentation method that yield acceptable results for every medical image. Nuclear imaging is the selective image modality for analyzing brain related neurological disorders. The major neurological alteration is accurate on shape, volume and tissue distribution in brain anatomy. PET scan images are exclusively used to identify the cause of brain neurological disorder.

Clustering is a major unsupervised learning technique. Fuzzy C- Means clustering is a well known soft segmentation method and it suitable for medical image segmentation than the crisp one. But, this conventional algorithm is calculated by iteratively minimizing the distance between the pixels and to the cluster centers. Spatial relationship of neighboring pixel is an aid of image segmentation. These neighboring pixels are highly correlated the same feature data. In spatial domain the membership of the neighbor centered are specified to obtain the cluster distribution statistics. Based on this statistics to calculate the weighting function and applied into the membership function [1]. For color image segmentation, Jafar et.al used HSV model for decomposition of color image and then FCM is applied separately on each component of it. The summation of the membership function in the neighborhood of each pixel provides the information of spatial function [2]. Yong yang handles small and large amount of noise in image segmentation to adjust a penalty coefficient using FCM algorithm [3, 4].

Alzheimer's Disease is a neurodegenerative disorder associated with memory loss. AD is often associated with reduced use of glucose in brain areas important in memory, learning and problem solving. In AD related research the volumetric analysis of hippocampus is one of the first regions of the brain to suffer memory problems. It acts as a memory indexer. It is used to consolidate the information from short term memory to long term memory and retrieving them when necessary. Human anatomy consists of two hippocampi, one in each side of the brain. But, the segmentation and identification of the hippocampus are highly complicated and more time consuming.

The organization of the paper is as follows. In section 2, the detail related work in this field is described. In section 3, spatial fuzzy c-means is explained. The proposed model is introduced in section 4. The experimental comparisons are presented in section 5. Finally, section 6, conclusion of the paper work is noted.

## 2. Related Work

Four various PET image segmentation methodologies such as thresholding method, variational approaches, stochastic modelling-based techniques and learning methods were selected in [5, 6]. Here, threshold value ($T$) is selected to separate the lesion foreground from a noisy background and PET image voxels are converted into standard uptake values. A logarithmic relationship between tumor volume and the optimal threshold using manually delineated CT data in a cohort of lung cancer patients is discussed by Biehl et al [7]. This studies have defined the optimal threshold selection is dependent on tumour volume. Wanlin Zhu discussed parameters assessment based iterative clustering method for segmentation of PET scan image datasets [8]. Elementary information retrieval technique is used for automatic detection of Alzheimer's disease using PET scan image. Residual vector analysis is applied to search a database of PET scans. Active voxels are the strongest kinetic activity and density frequencies of multidimensional data at different frames are identified by pre clustering approach. Early frame is a significant evidence of noise. But the final clustering is most efficient because the pre clustering produces fewer but larger clusters [9, 10]. Iterative thresholding method is used for the segmentation of metastatic volumes in PET [11].

## 3. Spatial Fuzzy C-Means Clustering

Clustering is used to classify items into identical groups in the process of data mining. It also exploits segmentation which is used for quick bird view for any kind of problem. K-Means is a well known partitioning method. Objects are classified as belonging to one of k groups, k chosen a priori [12]. Cluster membership is determined by calculating the centroid for each group and assigning each object to the group with the closest centroid. This approach minimizes the overall within-cluster dispersion by iterative reallocation of cluster members. Based on this, K-means segmentation of Alzheimer's Disease in PET scan datasets is implemented [13]. This produces the proven results for PET scan datasets using K-Means clustering. But in same datasets, if different structures exist, it has often found to fail. In general, the fuzzy c-means algorithm is assigned the pixels to fuzzy clusters without any label. Hard clustering methods are used to group pixels to belong exclusively one cluster. But, FCM allows a pixel in more than one cluster depends on the degrees of membership. Summation of membership of each data points in the given datasets should be equal to each other. Let $X = \{x_1, x_2, x_3 ..., x_n\}$ be the set of data points and $C = \{c_1, c_2, c_3 ..., c_n\}$ be the set of centers. The following equations 1 and 2 explain the membership and cluster center updation for each iteration.

$$\mu_{ij} = \frac{1}{\sum_{k=1}^{c} (d_{ij}/d_{ik})^{(2/m-1)}} \quad - (1)$$

$$c_j = \sum_{i=1}^{n} \left( \frac{((\mu_{ij})^m \, x_i)}{(\mu_{ij})^m} \right) \quad - (2)$$

where,

    **d$_{ij}$** represents the distance between ith data and jth cluster center.
    **c** represents the number of cluster
    **m** is the fuzziness index
    **µ$_{ij}$** represents the membership of ith data to jth cluster center.
    **n** is the number of data points.
    **c$_j$** represents the jth cluster center

Segmentation of Alzheimer's Disease in PET scan datasets using FCM is developed in [14]. Currently many researchers are included the spatial information in to the basic FCM algorithm to refine the segmentation result in medical images [15, 16]. Spatial fuzzy c-means, the spatial domain are updated based on the membership values of neighboring pixels. Huynh proposed an algorithm to utilize both given pixel attributes and the spatial local information which is weighted correspondingly to neighbor elements based on their distance attributes [17]. Dao et.al is proposed an algorithm using a kernel-induced distance metric and a spatial penalty on the membership functions [18]. This paper provides a description of the brain neuro degenerative disorder PET scan datasets and the implementation results of the SFCM.

## 4. Proposed Model

Fig. 1 illustrates the flowchart of the proposed PET SFCM segmentation for the brain neuro degenerative disorder.

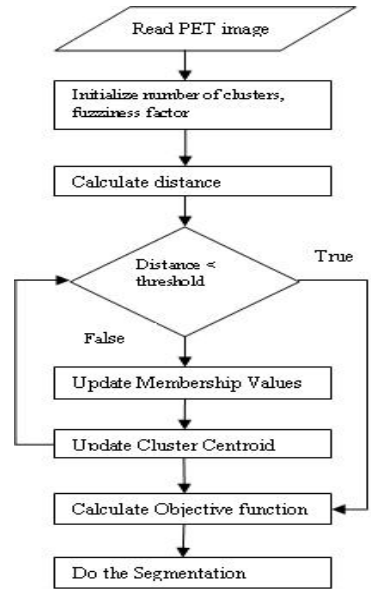

Fig.1: Flowchart of the proposed PET-SFCM

PET – SFCM Segmentation algorithm

    (1)  Read data from PET scan image
    (2)  Randomly select the number of clusters
    (3)  Initialize the fuzziness factor
    (4)  For i=1:max_iter
           i) Calculate the distance between
              pixel and centroid
           ii) Calculate the membership values
       End for
    (5)  If  distance < threshold
           i) Update the membership values
           ii) Update cluster centroid
      Else
         Determine objective function
      End if
    (6)  Do the Segmentation

A PET image from male patient age above 45 consists of nearly 266 samples are used in this implementation. Here, randomly, the number of cluster is selected as 5 and the performance iteration is varied. The initialization of fuzziness factor is 2. Each iteration, distance between each pixel and the cluster centroid is calculated. If distance is less than the threshold value, objective function is determined. Otherwise update the membership and cluster centroid values. Finally, segmentation is done according to the calculated threshold value.

## 5. Experimental Results

Spatial fuzzy c-means algorithm is implemented in MATLAB environment. The primary reason for the selection of MATLAB is significant amount of data available in that format and due to the increasing popularity of this language there is an extensive quantity of applications available. The whole datasets consists of two groups: one patient of probable Alzheimer's disease having 266 samples and one individual is normal. The below fig. 2(a) is shown the normal brain PET images and fig. 2(b) is shown the focally decreased glucose metabolism in alzheimer's disease.

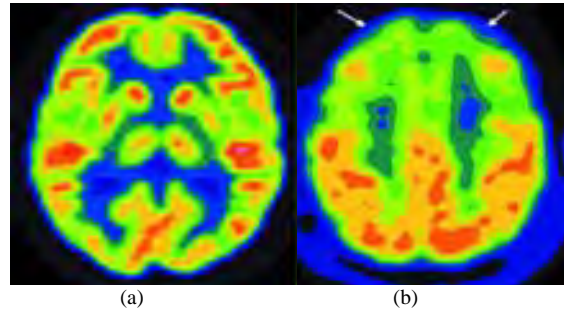

(a)                    (b)

Fig.2: (a) FDG-PET images show normal brain (b) Focally decreased glucose metabolism in Alzheimer disease

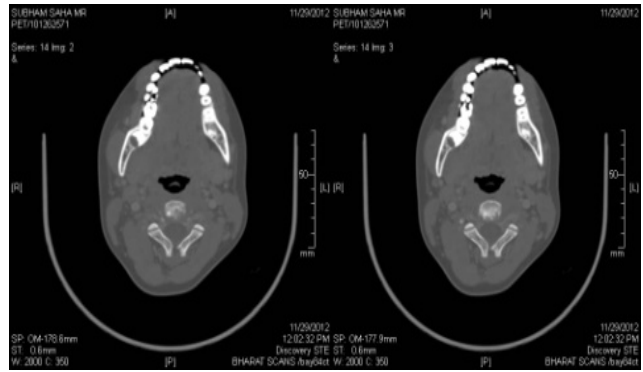

Fig.3:  FDG-PET images of 45 year male patient (Series 2/266 and 3/266)

A male patient of age above 45 years FDG-PET brain image samples as shown in fig.3. The below Fig. 4(b) shows the segmented image obtained from K-Means clustering algorithm with the number of clusters are randomly chosen as 5.  The comparison results of PET segmented image from Fuzzy C-Means clustering as shown in fig.5. The given FCM consists of 5 clusters randomly chosen as [25, 50, 75, 100 and 125] with iteration 25.

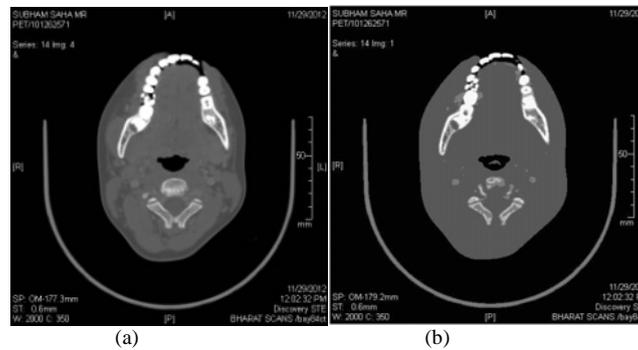

(a)                    (b)

Fig.4: (a) FDG-PET images of 45 year male patient (b) Segmented image obtained from K-Means clustering

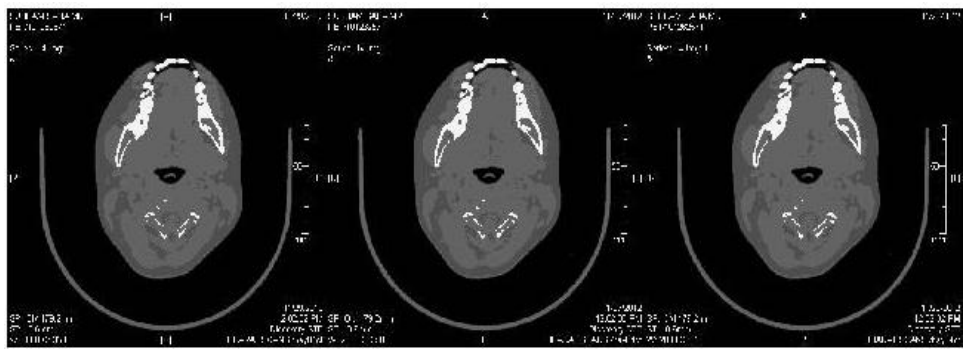

Fig.5: Segmented image obtained from FCM clustering with iteration 25

The proposed model PET-SFCM resultant image is shown in fig.6 and fig.7. Here the number of iterations is fixed as 25 and 75. Table 1 expressed the experimental parameters of various algorithms chosen in this paper.

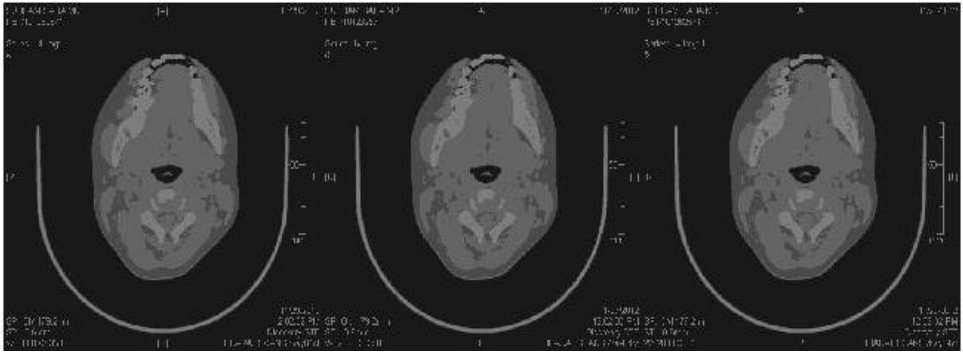

Fig.6: Segmented image obtained from proposed PET –SFCM clustering with iteration 25

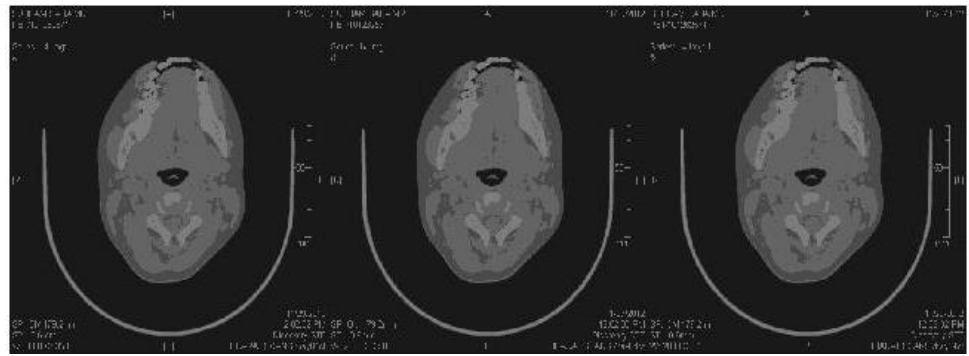

Fig.7: Segmented image obtained from proposed PET –SFCM clustering with iteration 75.

Table.1 Experimental Parameters

| Algorithm | No. of Clusters | Exponent | Maximum no. of iterations | Condition |
|-----------|-----------------|----------|---------------------------|-----------|
| K-Means | 5 | - | - | Distance < |
| FCM | 5 | 2 | 25 | Threshold |
| PET-SFCM | 3 | 2 | 25 and 75 | |

### 6. Conclusion

The nuclear imaging techniques have been prominently used for the clinical purpose such that the anatomy and the functional activities can be visualized. PET scan is a bio medical nuclear imaging techniques provide a solution for abnormal cells. The segmented image provides the clear picture about the affected portions. The proposed PET-SFCM is incorporated the spatial neighborhood information with traditional FCM. This algorithm is tested on huge data collection of patients with brain neuro degenerative disorder such as Alzheimer's disease. It has demonstrated its effectiveness by testing it for real world patient data sets of male patient age above 45 consists of nearly 266 samples. Experimental results are compared with conventional FCM and K-Means clustering algorithm. The performance of the SFCM provides satisfactory results compared with other two algorithms. In future, to calculate objective based quality assessment that could analyze images and report their quality without human involvement.